\documentclass[]{spie}  

\usepackage{amsmath,amsfonts,amssymb}
\usepackage{graphicx}
\usepackage[colorlinks=true, allcolors=blue]{hyperref}
\usepackage{subcaption}
\usepackage{booktabs}
\usepackage{xcolor}

\title{The Role of Noisy Data in Improving CNN Robustness for Image Classification}

\author[a]{Oscar H. Ram\'irez-Agudelo}
\author[a]{Nicoleta Gorea}
\author[a]{Aliza Reif}
\author[a]{Lorenzo Bonasera}
\author[a]{Michael Karl}

\affil[a]{German Aerospace Center (DLR), Institute for AI Safety and Security, Rathausallee 12, 53757 Sankt Augustin, Germany}

\authorinfo{Further author information: (Send correspondence to O.H.R-A.)\\O.H.R-A.: E-mail: Oscar.RamirezAgudelo@dlr.de}

\begin{document} 
\maketitle

\begin{abstract}
Data quality plays a central role in the performance and robustness of convolutional neural networks (CNNs) for image classification. While high-quality data is often preferred for training, real-world inputs are frequently affected by noise and other distortions. This paper investigates the effect of deliberately introducing controlled noise into the training data to improve model robustness. Using the CIFAR-10 dataset, we evaluate the impact of three common corruptions, namely Gaussian noise, Salt-and-Pepper noise, and Gaussian blur at varying intensities and training set pollution levels. Experiments using a Resnet-18 model reveal that incorporating just 10\% noisy data during training is sufficient to significantly reduce test loss and enhance accuracy under fully corrupted test conditions, with minimal impact on clean-data performance. These findings suggest that strategic exposure to noise can act as a simple yet effective regularizer, offering a practical trade-off between traditional data cleanliness and real-world resilience. 
\end{abstract}

\keywords{deep learning, CNNs, data quality, CIFAR-10, noise injection, image classification, model robustness}

\section{Introduction}\label{sec:introduction} 

In recent years, convolutional neural networks (CNNs) have formed the backbone of numerous computer vision applications, achieving state-of-the-art results in image classification, object detection, and segmentation tasks \cite{krizhevsky2012imagenet, he2016deep, lecun2015deep, rawat2017deep}. These advancements have been largely driven by increased computational power, larger annotated datasets, and improved network architectures. However, such success is critically dependent on the quality of the data used during training and evaluation. Data quality, described by attributes such as accuracy, consistency, completeness, and representativeness, strongly influences model performance, generalizability, and robustness in practical deployment scenarios \cite{bengio2013representation, sun2017revisiting, recht2019imagenet}.

Despite the availability of high-quality benchmark datasets like ImageNet \cite{deng2009imagenet} and medium-scale benchmarks such as CIFAR \cite{krizhevsky2009learning}, real-world image data often contains imperfections. It is susceptible to various degradations, including sensor noise, poor lighting, motion blur, occlusion, and compression artifacts \cite{dodge2016understanding, geirhos2018generalisation, karahan2016image}. These distortions can substantially impair model performance, particularly when the training data does not reflect such variability. The resulting domain shift between clean training data and noisy real-world inputs poses a significant challenge to robustness \cite{hendrycks2019benchmarking, ford2019adversarial, taori2020measuring}.

To address this challenge, researchers should focus on both data quality management and model robustness. While conventional best practices emphasize curating clean and balanced datasets, recent work has shown that strategically incorporating degraded or noisy data during training can improve robustness to real-world noise \cite{tradeoffs_clean_noisy,shorten2019survey, zhang2017mixup, xie2020adversarial}. Instead of removing all noise entirely, exposing models to controlled levels of degradation can serve as a form of regularization, enhancing their ability to generalize to unseen conditions \cite{liu2020simple, rusak2020simple}.

In this study, we investigate how different types of image degradation, specifically \emph{Gaussian noise}, \emph{Salt-and-Pepper noise}, and \emph{Gaussian blur} affect the robustness of CNNs in image classification tasks. Building on prior research into noise robustness and data augmentation, we conduct a series of experiments using the CIFAR-10 dataset, adding different intensities of Gaussian noise, Salt-and-Pepper noise, and Gaussian blur to evaluate their impact on model performance \cite{krizhevsky2009learning, madry2018towards}. We train the resulting models on both clean and noisy images to evaluate how noise exposure during training influences classification performance on both clean and degraded test sets. Performance is assessed using top-1 classification accuracy, with 10\% of the training set held out for validation (see more details in Sect.~\ref{sec:methodology}).
The contributions of this work are threefold:
\begin{itemize}
\item We conduct an empirical analysis to evaluate the effect of noisy data inclusion on CNN robustness, adding to recent efforts in data-centric deep learning \cite{northcutt2021pervasive}.
\item We propose a simple yet effective noise-augmented training strategy that improves generalization to noisy test conditions while maintaining strong performance on clean data \cite{zhang2021understanding}.
\item We discuss trade-offs between classical definitions of high data quality and the benefits of incorporating noise to promote real-world adaptability \cite{recht2019imagenet, beery2018recognition}.
\end{itemize}

The findings underscore the need to reconsider rigid definitions of data quality. We advocate for a broader perspective that values both cleanliness and representativeness of operational environments. This shift carries important implications for building machine learning systems that operate reliably under unpredictable and adverse real-world conditions.

Finally, we structure this paper as follows. Section~\ref{sec:sota} reviews related literature. Section~\ref{sec:methodology} outlines the methodology used in this study. Section~\ref{sec:results} presents the results, and Section~\ref{sec:discussion} discusses them in detail. Section~\ref{sec:conclusion} concludes the paper. Appendix~\ref{sec:cifar-10} reports the baseline performance of the adopted model prior to the experiments described in this work.


\section{RELATED WORKS}\label{sec:sota}

The importance of \textit{data quality} in training robust machine learning models has been extensively studied. Datasets such as ImageNet \cite{deng2009imagenet} and CIFAR-10  \cite{krizhevsky2009learning} have been foundational for advancing deep learning in image classification. These datasets are carefully curated to ensure consistency and representativeness, which are critical for reliable model performance \cite{dataset_curation_ml, impact_data_quality, annotation_consistency}.

Despite these benchmarks, real-world images often suffer from degradations including noise, blur, and compression artifacts, which can negatively impact model accuracy when models are trained solely on clean data \cite{noise_robustness_cnns, image_degradation_impact}. Addressing this, Hendrycks and Dietterich \cite{hendrycks2019benchmarking} benchmarked the robustness of neural networks against a wide range of common corruptions, demonstrating the vulnerability of standard CNNs to noise and other perturbations.

The work by De Vries et al. \cite{devries2017improvedregularizationconvolutionalneural} introduces Cutout, a regularization technique that randomly masks out square regions of input images during training. While Cutout is not explicitly designed for robustness to noise or corruptions, it improves generalization and has been shown to enhance accuracy on clean test data. Additionally, data augmentation has emerged as a practical method to improve robustness by introducing variability during training \cite{shorten2019survey, data_augmentation_robustness, tradeoffs_clean_noisy}. For example, Lopes et al. \cite{tradeoffs_clean_noisy} propose Patch Gaussian, which injects Gaussian noise into random localized regions of input images. This method improves robustness to common corruptions without sacrificing and sometimes even improving accuracy on clean data.

Beyond generic augmentation, adversarial training, as introduced by Goodfellow et al. \cite{goodfellow2015explaining}, involves training models on deliberately perturbed examples to increase robustness against worst-case input alterations. Although adversarial robustness focuses on targeted perturbations, it shares the goal of improving performance under degraded input conditions \cite{fawzi2016robustness}.

Techniques such as random cropping, flipping, and synthetic noise injection are widely used to enhance generalization \cite{hendrycks2018using}. In particular, noise injection during training has been shown to encourage models to learn more invariant features, improving resilience to noisy inputs \cite{hendrycks2018using}. Techniques such as mixup \cite{zhang2018mixup} blend clean and noisy samples to achieve better generalization and robustness.

Additional strategies to balance noise include curriculum learning \cite{curriculum_learning_noisy}, noise-aware loss functions \cite{noise_aware_loss}, and adaptive noise filtering during training \cite{adaptive_noise_filtering}. Domain adaptation techniques have also been applied to address shifts caused by image quality variations \cite{domain_adaptation_methods}.

This work aims to complement these efforts by empirically analysing the effect of incorporating different type of noises such as Gaussian noise, Salt-and-Pepper noise, and Gaussian blur into CIFAR-10 benchmark dataset as a test bed. By evaluating classification accuracy on both clean and noisy test sets, we aim to clarify how \textit{data quality}, viewed as a spectrum from clean to noisy, influences model robustness in practical settings.


\section{METHODOLOGY}\label{sec:methodology}

This section describes the methodology adopted to investigate the impact of training data quality degradation on the robustness of convolutional neural networks (CNNs) for image classification by using the CIFAR-10 dataset \cite{krizhevsky2009learning}. In Section~\ref{sec:data_processing} details about the data, architecture, and training are provided. Subsequently, in Section~\ref{sec:type_of_noise} information about the type of noise and values selected are presented. Finally, the metrics used to evaluate the models are introduced in Section~\ref{sec:metrics}.

\subsection{Model design and training overview}\label{sec:data_processing}

This section provides a comprehensive overview of the model design and training methodology employed in this work. 

\subsubsection{Dataset}\label{sec:dataset}

The CIFAR-10 dataset contains 60,000 color images, each with a resolution of 32×32 pixels and three RGB channels. These images are evenly distributed across 10 classes: plane, car, bird, cat, deer, dog, frog, horse, ship, truck. This widely studied dataset offers excellent opportunities for comprehensive investigation and benchmarking \cite{tradeoffs_clean_noisy,krizhevsky2009learning,krizhevsky2012imagenet,he2016deep}. Due to its manageable size and complexity, CIFAR-10 is frequently used to evaluate novel machine learning architectures, optimization techniques, and regularization strategies. Despite its simplicity compared to large-scale datasets like ImageNet, it remains a valuable testbed for rapid prototyping and ablation studies, particularly in convolutional neural networks (CNNs) and more recent deep learning models. Furthermore, CIFAR-10 continues to serve as a baseline dataset in exploring model robustness, transfer learning, and self-supervised learning paradigms.

\subsubsection{Model Architecture}\label{sec:architecture}

For this work, the Resnet-18 architecture implemented in the Github repository \url{https://github.com/kuangliu/pytorch-cifar} by \emph{kuangliu} is adopted. Appendix~\ref{sec:cifar-10} shows the training performance achieved by adopting this repository. An accuracy of approximately 93\% is being retrieved, which is also reported in the open-source repository. This architecture serves as the backbone for our evaluations. To ensure fair comparisons, consistent training hyperparameters across all conditions are maintained. The network is trained using stochastic gradient descent (SGD) with a fixed learning rate $\alpha = 0.01$ and batch size of 128 and 100 for the training and testing set, respectively and total of 100 epochs. We select the best-performing model based on validation loss.

\subsubsection{Training}\label{sec:training}

For the experimental setup, we split 45000 images for training (90\% of the original training set), 5000 images for validation (10\% of the training set), and kept the test set unchanged at 10,000 images. We apply standard preprocessing, including normalization and random cropping, consistently across all experiments, following the implementation provided by \emph{kuangliu}. To ensure robust results, we run each experiment 10 times per noise level and calculate the mean and standard deviation. This repetition mitigates the effects of randomness inherent in training, such as weight initialization, data shuffling, and stochastic optimization. By reducing variance and increasing consistency, this approach yields more reliable and statistically sound performance estimates, while also improving reproducibility by ensuring that results reflect genuine model behavior rather than random fluctuations.

\subsection{Noise Injection}\label{sec:type_of_noise}

We introduce three different types of noise: Gaussian, Salt-and-Pepper, and Gaussian blur. To analyze the impact of noise severity on training, we create multiple training subsets with varying percentages of noisy images, corresponding to the following pollution levels: 0\% (clean), 5\%, 10\%, 15\%, 20\%, 25\%, 50\%, 75\%, and 100\% (fully noisy). For each pollution level (9 levels), the corresponding fraction of training samples is randomly selected and corrupted by adding noise, while the respective rest remains clean. In the following, we describe the characteristics of each noise type in detail. This approach models realistic scenarios where data quality varies within the training set, allowing us to systematically evaluate the effect of mixed-quality data on model robustness.

\subsubsection{Gaussian Noise}

Gaussian noises with zero mean and several levels of standard deviation ($\sigma$) are added to the training images to simulate quality degradation\footnote{The values of the levels are introduced in Section~\ref{sec:values}.}. Specifically, we focus on the effect of Gaussian noise introduced at varying levels to the training data, examining how different proportions of noisy samples influence model performance.

To assess the robustness of the proposed approach under varying levels of input perturbations, we inject additive Gaussian noise into the input data. The noise is sampled from a normal distribution $\mathcal{N}(0, \sigma^2)$, where $\sigma$ denotes the standard deviation controlling the noise intensity. This simulates naturally occurring sensor noise or environmental variability in real-world acquisition settings. The noisy input $I_{\text{noisy}}$ is generated as:
\begin{equation}
    I_{\text{noisy}} = I_{\text{clean}} + N_{\text{gaussian}}
\end{equation}
where $I_{\text{clean}}$ is the original input and $N_{\text{gaussian}} \sim \mathcal{N}(0, \sigma^2)$ is the zero-mean Gaussian noise.

The probability density function of the Gaussian distribution is given by:
\begin{equation}
    p(x) = \frac{1}{\sqrt{2\pi\sigma^2}} \exp\left( -\frac{(x - \mu)^2}{2\sigma^2} \right)
\end{equation}
with $\mu = 0$ and varying values of $\sigma$ to control the noise strength.

\subsubsection{Salt-and-Pepper Noise}

To evaluate the robustness of the proposed approach against impulsive noise, we study the introduction of Salt-and-Pepper noise to the input data. Salt-and-Pepper is a form of impulse noise where certain pixel values are randomly replaced with either the minimum (``pepper'') or maximum (``salt'') intensity values in the input range \cite{sontakke2015}. This simulates sensor malfunctions or bit errors in transmission.
We generate the noisy input $I_{\text{noisy}}$ as follows:

\begin{equation}
    I_{\text{noisy}}(i,j) = 
    \begin{cases}
        I_{\text{min}}, & \text{with probability } p_{\text{pepper}} \\
        I_{\text{max}}, & \text{with probability } p_{\text{salt}} \\
        I_{\text{clean}}(i,j), & \text{with probability } 1 - p_{\text{salt\_and\_pepper}}
    \end{cases}
\end{equation}
\noindent
where $I_{\text{clean}}(i,j)$ denotes the original clean input at pixel location $(i,j)$, and $p_{\text{salt\_and\_pepper}} = p_{\text{salt}} + p_{\text{pepper}}$ is the total noise density. $I_{\text{min}}$ and $I_{\text{max}}$ represent the minimum and maximum allowable intensities, respectively. In this work, $p_{\text{salt}}$ and $p_{\text{pepper}}$ contribute equally.

\subsubsection{Gaussian Blur}

In addition to noise injection, we apply Gaussian blur to evaluate the model's robustness against loss of fine-grained detail. Gaussian blur is a common image transformation that simulates sensor defocus or motion-induced blur by convoluting the image with a Gaussian kernel. This operation smooths the image, attenuating high-frequency components such as edges and textures.
Formally, the blurred image $I_{\text{blurred}}$ is obtained by the convolution:
\begin{equation}
    I_{\text{blurred}}(i,j) = (I_{\text{clean}} * G_{\sigma})(i,j),
\end{equation}
where $G_{\sigma}$ denotes a 2D Gaussian kernel with standard deviation $\sigma$, defined as:
\begin{equation}
    G_{\sigma}(x, y) = \frac{1}{2\pi\sigma^2} \exp\left( -\frac{x^2 + y^2}{2\sigma^2} \right)
\end{equation}
The kernel is centered around $(x, y) = (0, 0)$, and the convolution is applied across all pixels in the input.

\subsubsection{Values}\label{sec:values}
\begin{figure}[htbp]
    \centering

    \begin{subfigure}[b]{0.7\textwidth}
        \centering
        \caption{Gaussian noise}
        \includegraphics[width=\textwidth]{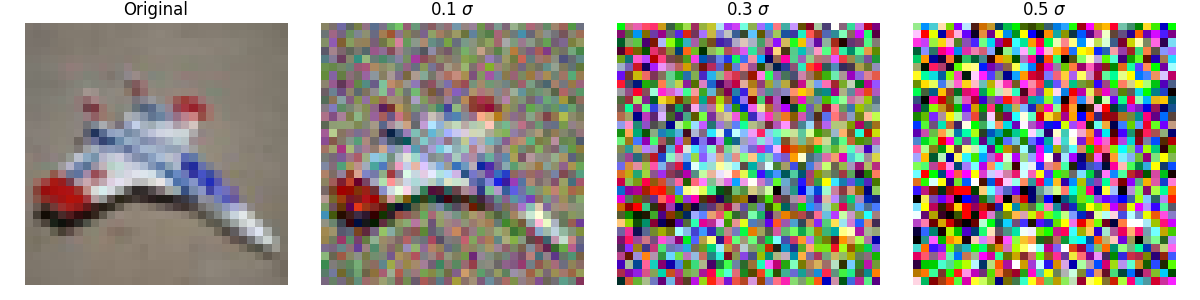}
        \label{fig:cifar10_gaussian}
    \end{subfigure}
    \hfill
    \begin{subfigure}[b]{0.7\textwidth}
        \centering
        \caption{Salt-and-Pepper noise}
        \includegraphics[width=\textwidth]{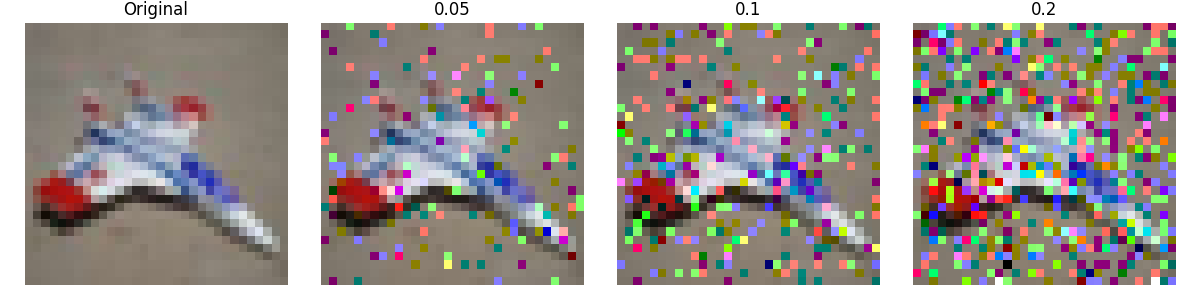}
        \label{fig:cifar10_salt_pepper}
    \end{subfigure}
    \hfill
    \begin{subfigure}[b]{0.7\textwidth}
        \centering
        \caption{Gaussian blur}
        \includegraphics[width=\textwidth]{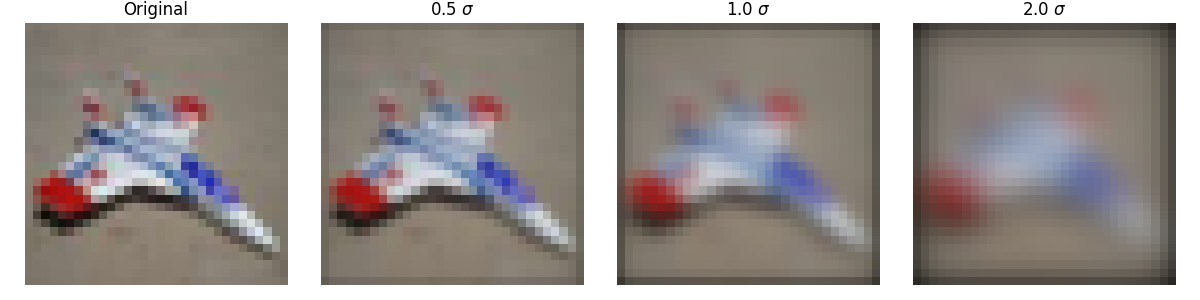}
        \label{fig:cifar10_blur}
    \end{subfigure}

    \caption{Impact of different types and intensities of image degradation on CIFAR-10.}
    \label{fig:cifar10_noise_comparison}
\end{figure}

To systematically evaluate the robustness of our method under varying data corruptions, we consider three severity levels (mild, moderate, and strong) based on the corresponding parameter values for each type of noise. Intensity values for the Salt-and-Pepper noise are derived by the work of Wan et al. \cite{Wan2024}. Instead, Gaussian blur the values are adapted from the work Yoshihara et al. \cite{Yoshihara2023}. Here, the moderate and strong levels are set to half of those used by the author\footnote{By doing so, we can sufficiently demonstrate our method.}. In the case of Gaussian noise, values are decided through preliminary experiments, based on comparable perceived degradation in relation with the other pollution types.
An overview of the parameter settings for each level is provided in Table~\ref{tab:noise_blur_levels}. Figure~\ref{fig:cifar10_noise_comparison} shows as an example the impact of the respective parameter settings for the 
class plane.

\begin{table}[h]
\renewcommand{\arraystretch}{1.25}
\centering
\caption{Noise and level of pollution categorized as mild, moderate, and strong, with associated parameters. For the Salt-and-Pepper noise, $p_{\text{salt}}$ and $p_{\text{pepper}}$ contribute equally. For example, when $p_{\text{salt\_and\_pepper}}=0.1$ (mild), corresponds to $p_{\text{salt}}=p_{\text{pepper}}=0.05$.}\vspace{0.3cm}

\begin{tabular}{ l c c c }
\toprule
\textbf{Type} & \textbf{Mild} & \textbf{Moderate} & \textbf{Strong} \\
\midrule
Gaussian noise ($\sigma$) & $0.1$ & $0.3$ & $0.5$ \\
Salt-and-Pepper noise  ($p_{\text{salt\_and\_pepper}}$)     &       0.05 & 0.1 & 0.2 \\ 
Gaussian blur ($\sigma_{\text{blur}}$) & $0.5$ & $1.0$ & $2.0$ \\ \bottomrule
\end{tabular}
\label{tab:noise_blur_levels}
\end{table}

\subsection{Evaluation Metrics and Analysis}~\label{sec:metrics}

Performance is primarily evaluated using classification accuracy on the clean test set and on test sets with varying the noise levels, to assess model robustness. Additionally, training and validation loss curves are analyzed to understand convergence behavior under different noise pollution levels. Results are aggregated over the 10 runs per pollution level to calculate mean and standard deviation for all metrics (see also Sect.~\ref{sec:training}). The main focus is to plot, compare, and assess the accuracy and loss as functions of the training data pollution percentage, revealing how incorporating noisy data during training influences robustness against noisy test inputs.

\section{Results}\label{sec:results}
\subsection{Loss}\label{sec:results_loss}
This section describes the results for the three different noise types in terms of test loss.

\subsubsection{Gaussian Noise}\label{sec:results_loss_gaussian}
\begin{figure}[ht]
    \centering
    \begin{subfigure}[b]{0.48\linewidth}
        \centering
        \includegraphics[width=\linewidth]{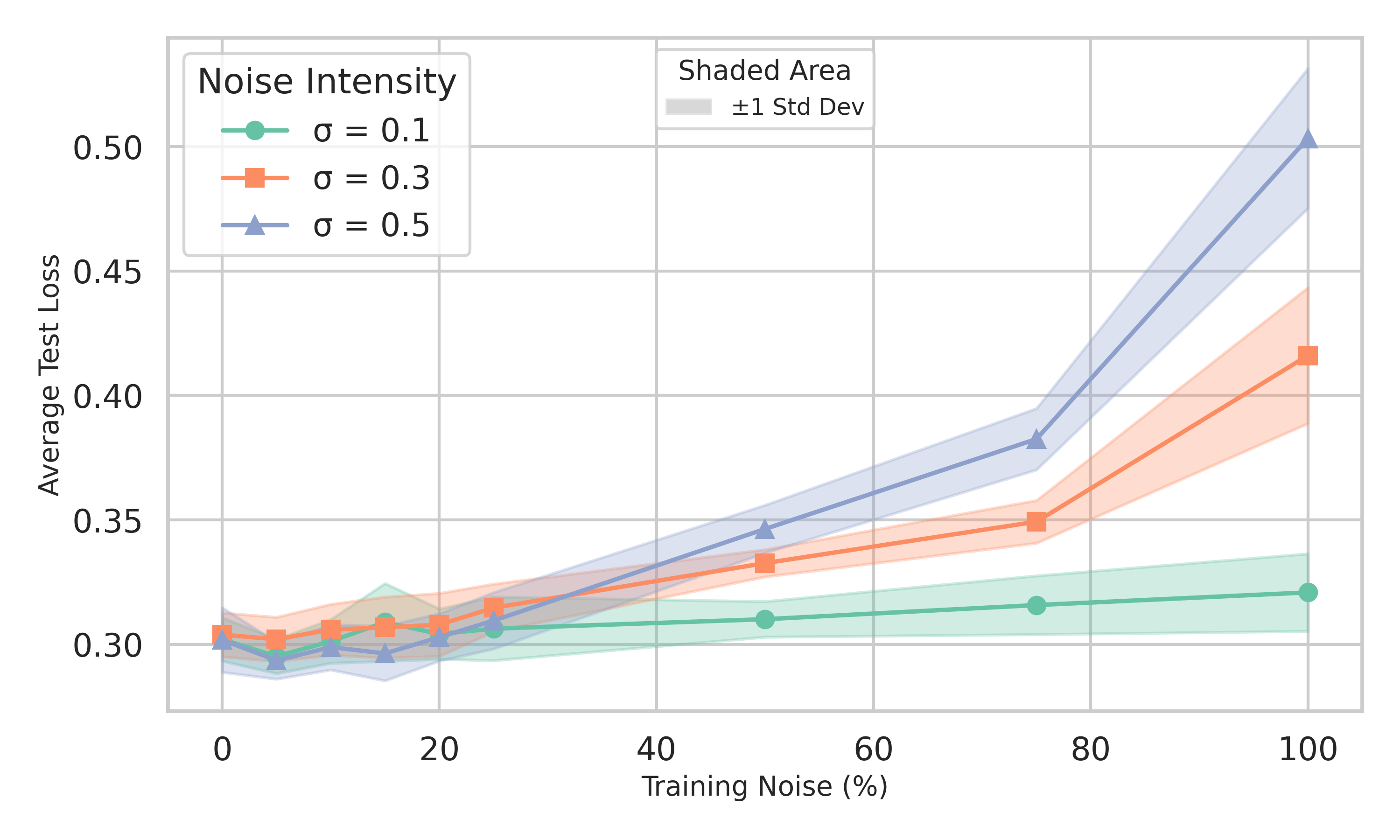}
        \caption{Loss of the three different intensities of Gaussian noise tested on the clean test set.}
        \label{fig:loss_gaussian_clean}
    \end{subfigure}
    \hfill
    \begin{subfigure}[b]{0.48\linewidth}
        \centering
        \includegraphics[width=\linewidth]{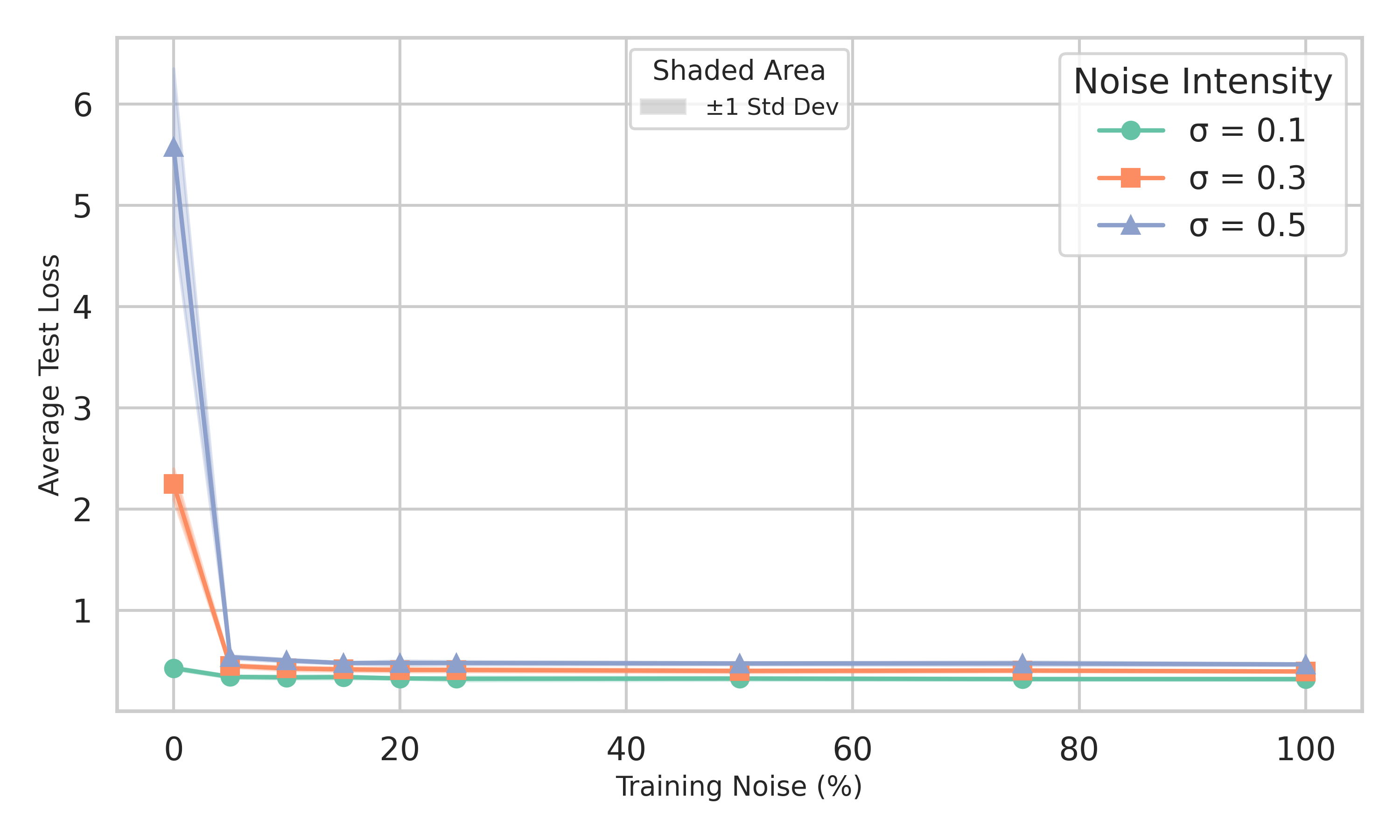}
        \caption{Loss of the three different intensities of Gaussian noise tested on the maximum noisy test set.}
        \label{fig:loss_gaussian_noisy}
    \end{subfigure}
    \caption{Average test loss across different Gaussian noise intensities and varying training noise percentages. Shaded areas indicate $\pm 1$ standard deviation over 10 runs.}
    \label{fig:loss_gaussian_dual}
\end{figure}

Figure~\ref{fig:loss_gaussian_dual} shows the average test loss of the model training on CIFAR-10 data with varying intensities of Gaussian noise $\sigma \in \{0.1, 0.3, 0.5\}$. Figure~\ref{fig:loss_gaussian_clean} shows the test performance for the three levels of Gaussian noise on the clean test set, revealing that models trained with small amounts of Gaussian noise ($\sigma = 0.1$) maintain the lowest average test loss even as training noise levels increase. In contrast, higher-intensity noise ($\sigma = 0.5$) begins to degrade performance, especially at higher training noise percentages.

Figure~\ref{fig:loss_gaussian_noisy} demonstrates the model’s performance on the corresponding fully corrupted test set. A small amount of noise augmentation during training (as low as $5\%$) leads to a substantial reduction in test loss, indicating that robustness improves with moderate exposure. However, training with excessively strong noise ($\sigma = 0.5$) results in higher test loss, suggesting diminishing returns or over-regularization. Overall, these plots highlight the trade-off between robustness and clean-data performance as a function of both noise intensity and training exposure. Moderate augmentation with low-intensity noise ($\sigma = 0.1$ or $0.3$) provides the best balance.  

\subsubsection{Salt-and-Pepper Noise}\label{sec:results_loss_salt_pepper}

\begin{figure}[ht]
    \centering
    \begin{subfigure}[b]{0.48\linewidth}
        \centering
        \includegraphics[width=\linewidth]{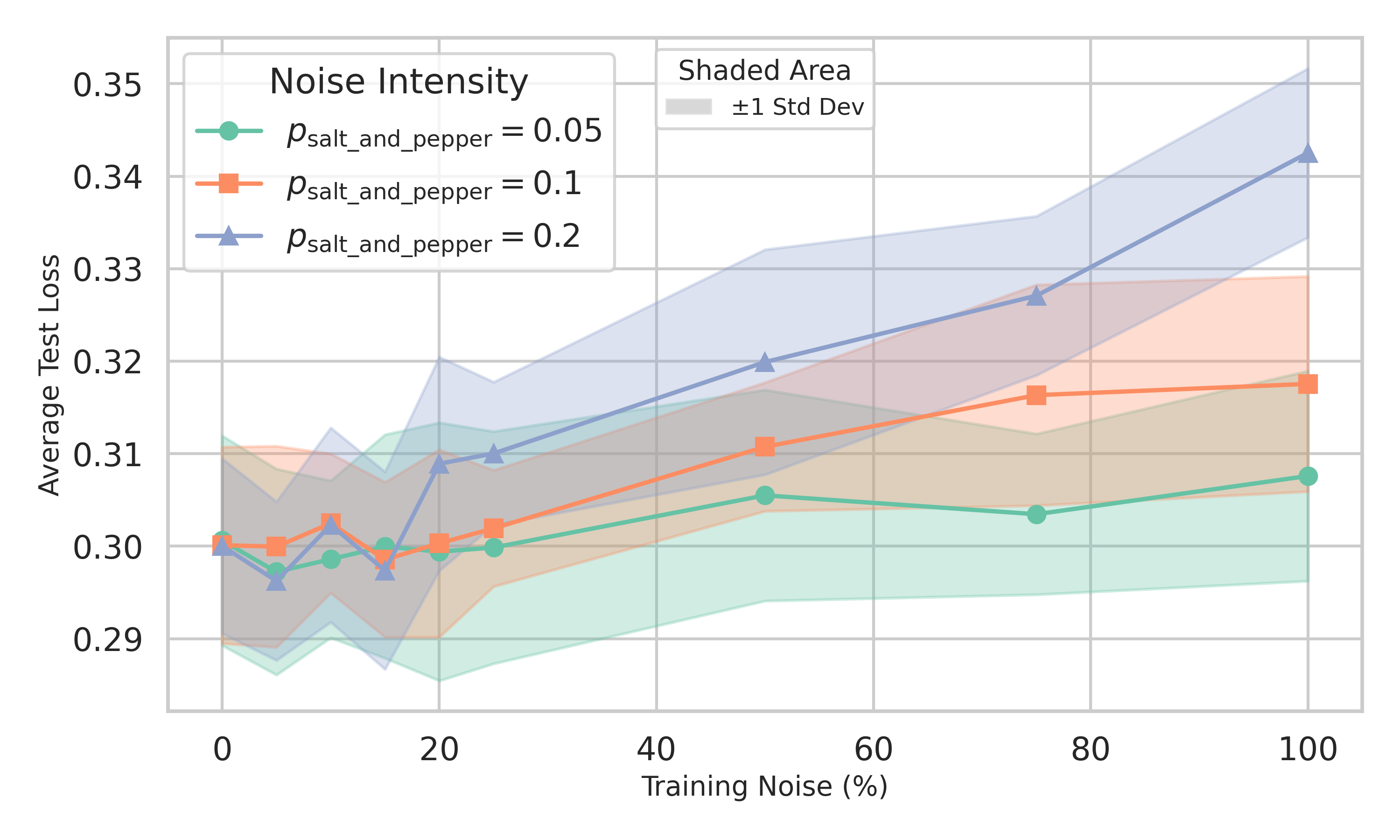}
        \caption{Loss of the three different intensities of Salt-and-Pepper noise tested on the clean test set.}
        \label{fig:loss_salt_pepper_clean}
    \end{subfigure}
    \hfill
    \begin{subfigure}[b]{0.48\linewidth}
        \centering
        \includegraphics[width=\linewidth]{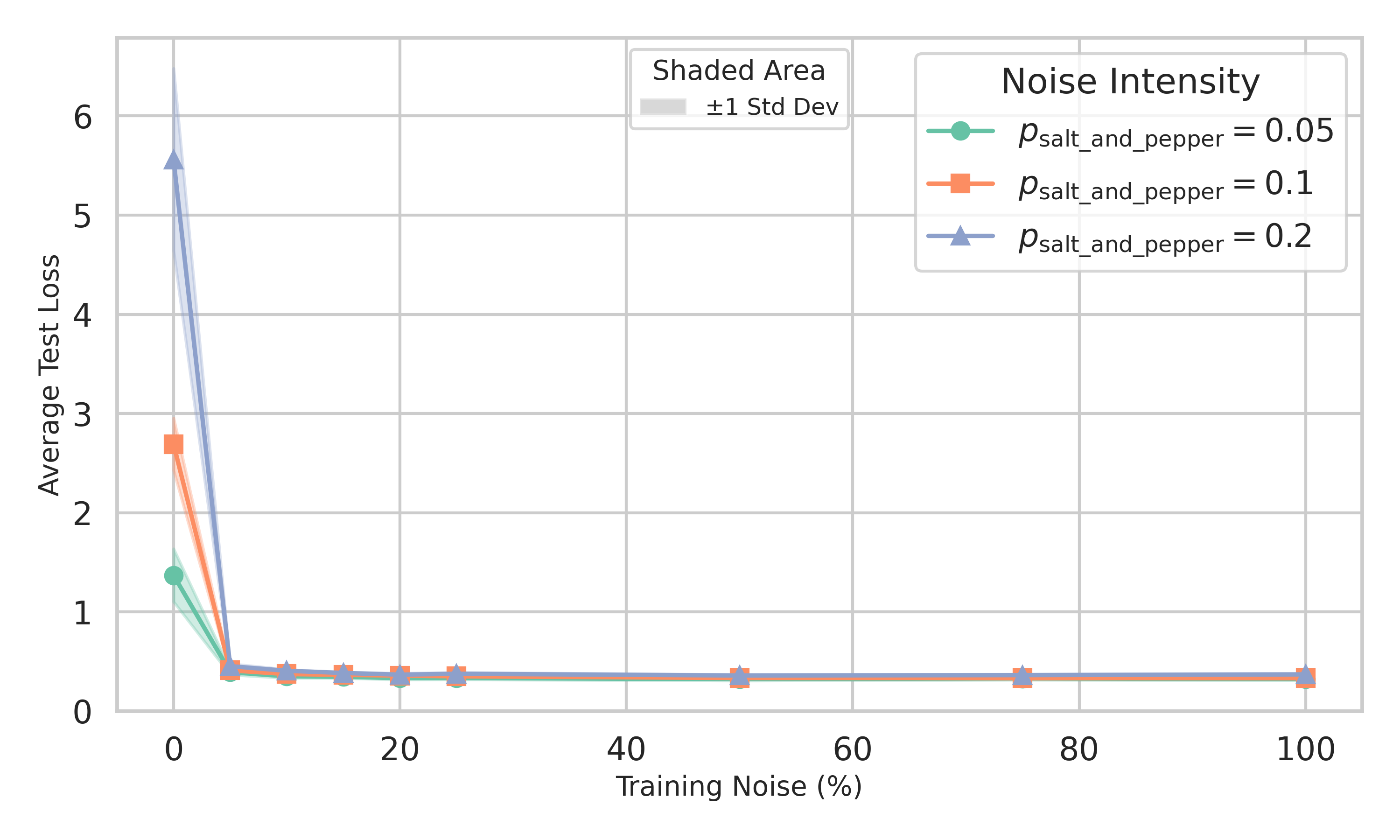}
        \caption{Loss of the three different intensities of Salt-and-Pepper noise tested on the maximum noisy test set.}
        \label{fig:loss_salt_pepper_noisy}
    \end{subfigure}
    \caption{Average test loss across different Salt-and-Pepper noise intensities and varying training noise percentages. Shaded areas indicate $\pm 1$ standard deviation over 10 runs.}
    \label{fig:loss_salt_pepper_dual}
\end{figure}

Figure~\ref{fig:loss_salt_pepper_dual} presents the average test loss of the model trained on CIFAR-10 data with varying levels of Salt-and-Pepper noise applied to the training set. Results are shown for three intensities \mbox{$p_{\text{salt\_and\_pepper}} \in \{0.05, 0.1, 0.2\}$} and are evaluated in the same manner as in Section~\ref{sec:results_loss_gaussian}: on clean test data (left) and fully corrupted test data (right). On clean test data, as shown in Figure~\ref{fig:loss_salt_pepper_clean}, increasing the proportion of noisy images in the training set generally results in higher test loss, with a more pronounced effect at higher noise levels. The model trained with $p_{\text{salt\_and\_pepper}} = 0.2$ consistently exhibits the highest loss, indicating a greater susceptibility to overfitting on noisy features.

Conversely, when evaluated on fully corrupted test data, as shown in Figure~\ref{fig:loss_salt_pepper_noisy}, the trend reverses: models trained without noise perform poorly, while those trained with any of the tested noise intensities achieve significantly lower test loss. The most notable improvement occurs when increasing training noise from 0\% to 10\%, after which performance levels off across all intensities. These results suggest that adding a small amount of noise to the training data acts as an effective regularizer, enhancing the model's robustness to real-world corruptions. Overall, the findings underscore the importance of aligning training conditions with deployment scenarios when robustness to noise is required.

\subsubsection{Gaussian Blur}\label{sec:results_loss_gaussian_blur}

\begin{figure}[ht]
    \centering
    \begin{subfigure}[b]{0.48\linewidth}
        \centering
        \includegraphics[width=\linewidth]{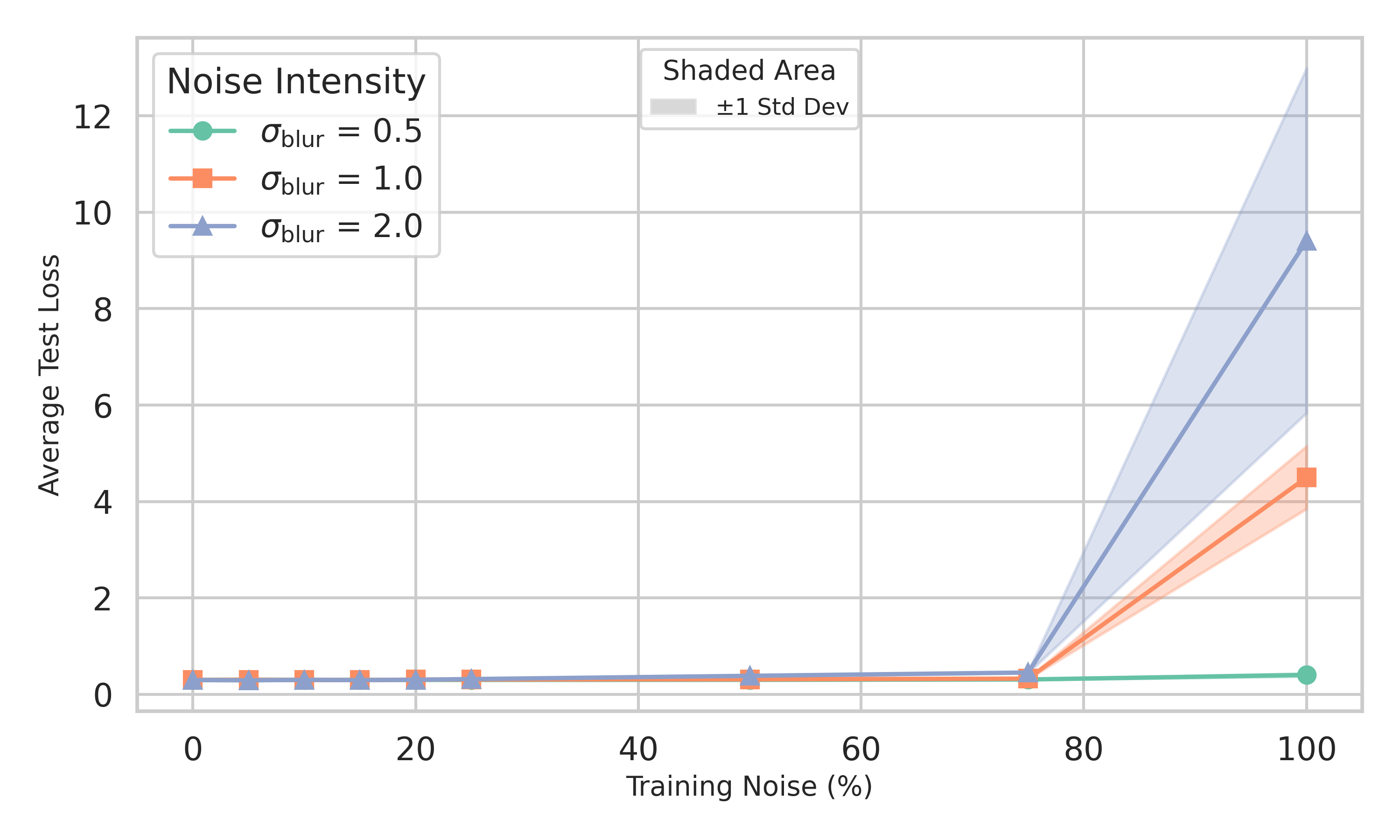}
        \caption{Loss of the three different intensities of Gaussian blur tested on the clean test set.}
        \label{fig:loss_gaussian_blur_clean}
    \end{subfigure}
    \hfill
    \begin{subfigure}[b]{0.48\linewidth}
        \centering
        \includegraphics[width=\linewidth]{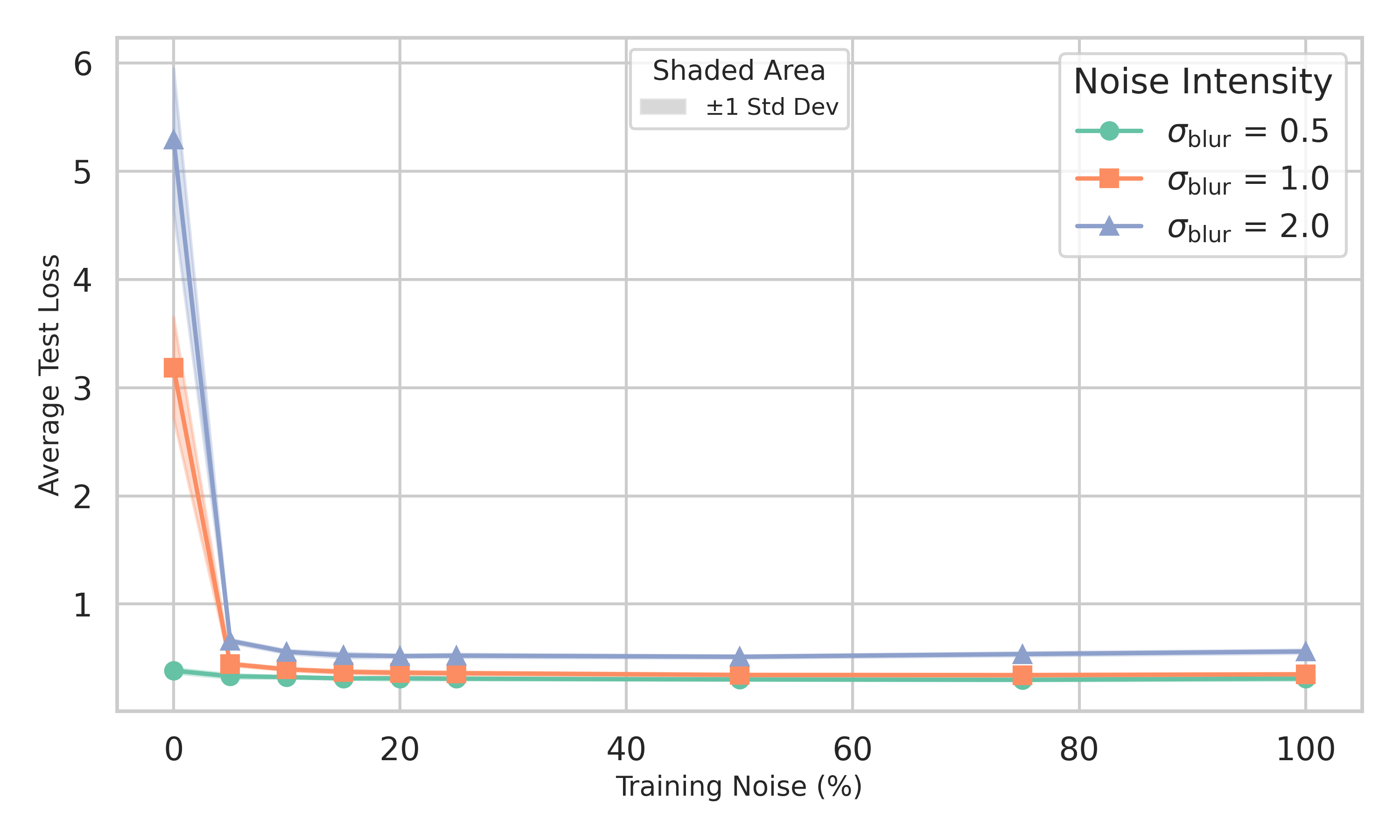}
        \caption{Loss of the three different intensities of Gaussian blur tested on the maximum noisy test set.}
        \label{fig:loss_gaussian_blur_noisy}
    \end{subfigure}
    \caption{Average test loss across different Gaussian blur intensities and varying training noise percentages. Shaded areas indicate $\pm 1$ standard deviation over 10 runs.}
    \label{fig:loss_gaussian_blur_dual}
\end{figure}

Figure~\ref{fig:loss_gaussian_blur_dual} presents the average test loss of the model trained on CIFAR-10 data with varying levels of Gaussian blur applied to the training set. The results are shown for three different noise intensities, $\sigma_{\text{blur}} \in \{0.5, 1.0, 2.0\}$, and are evaluated as described in the previous sections. On clean test data, as shown in Figure~\ref{fig:loss_gaussian_blur_clean}, increasing the proportion of noisy images in the training set generally leads to higher test loss, with the model trained with $\sigma_{\text{blur}} = 2.0$ consistently exhibiting the highest loss. This indicates its vulnerability to overfitting on blurred features.  

On the contrary, when evaluated on fully corrupted test set shown in Figure~\ref{fig:loss_gaussian_blur_noisy}, the trend is reversed: models trained with low or no noise perform poorly, while those trained with moderate noise intensities ($\sigma_{\text{blur}} = 1.0$ and $2.0$) achieve significantly lower test loss. The most substantial improvement occurs when the noise is of the level of 5\% in the training noise, after which the performance plateaus across all intensities. Similarly to the results of the previous sections, these results also indicate that adding a small degree of noise to the training data can act as a beneficial form of regularization, helping the model better withstand input degradation. This underscores the value of tailoring training conditions to reflect the types of noise and variability expected during deployment, especially when robustness is critical.

\subsection{Accuracy}\label{sec:results_acc}
This section describes the results for the three different noise types in terms of test accuracy.

\subsubsection{Gaussian Noise}\label{sec:results_acc_gaussian}
\begin{figure}[ht]
    \centering
    \begin{subfigure}[b]{0.48\linewidth}
        \centering
        \includegraphics[width=\linewidth]{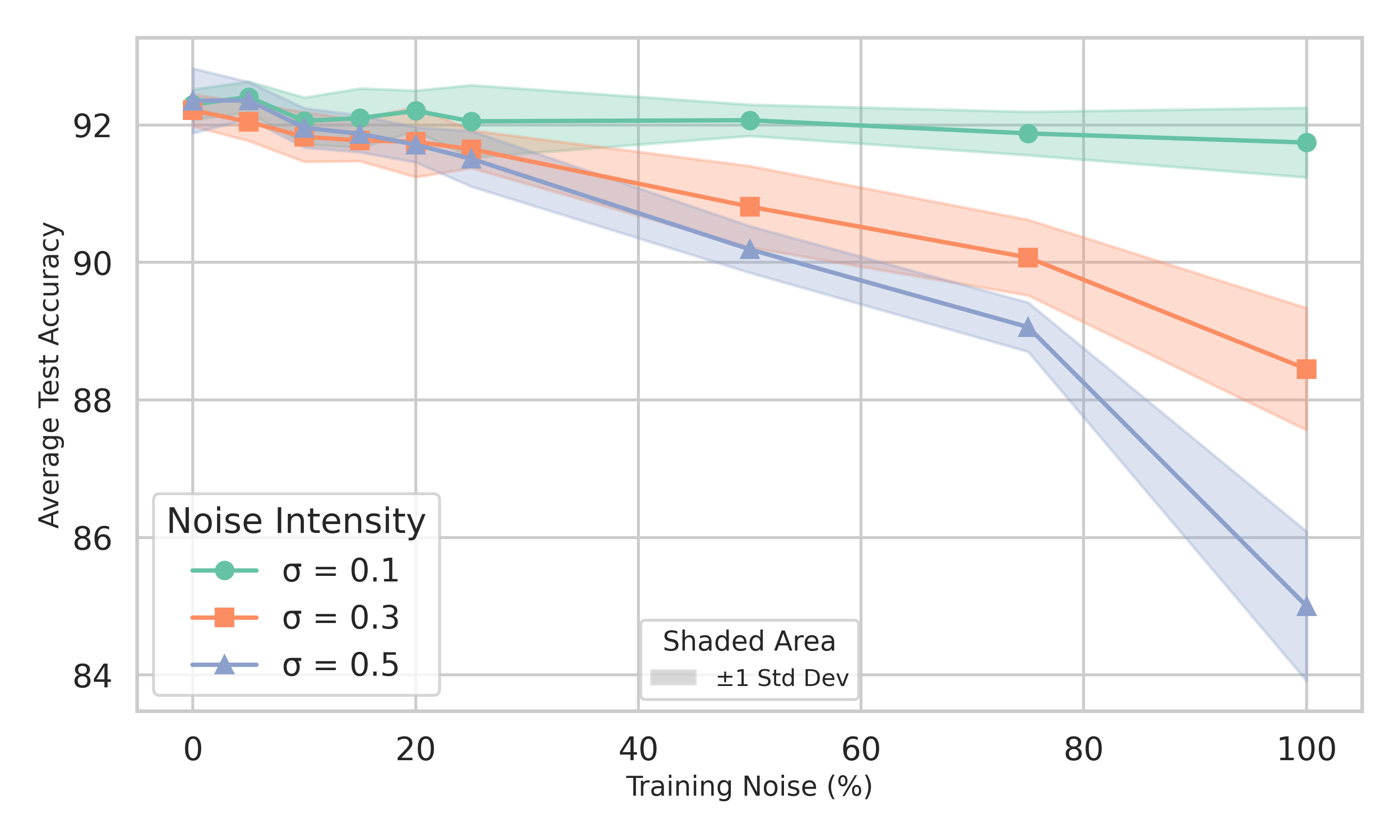}
        \caption{Effect of training noise on fully clean test performance.\\}
        \label{fig:acc_gaussian_clean}
    \end{subfigure}
    \hfill
    \begin{subfigure}[b]{0.48\linewidth}
        \centering
        \includegraphics[width=\linewidth]{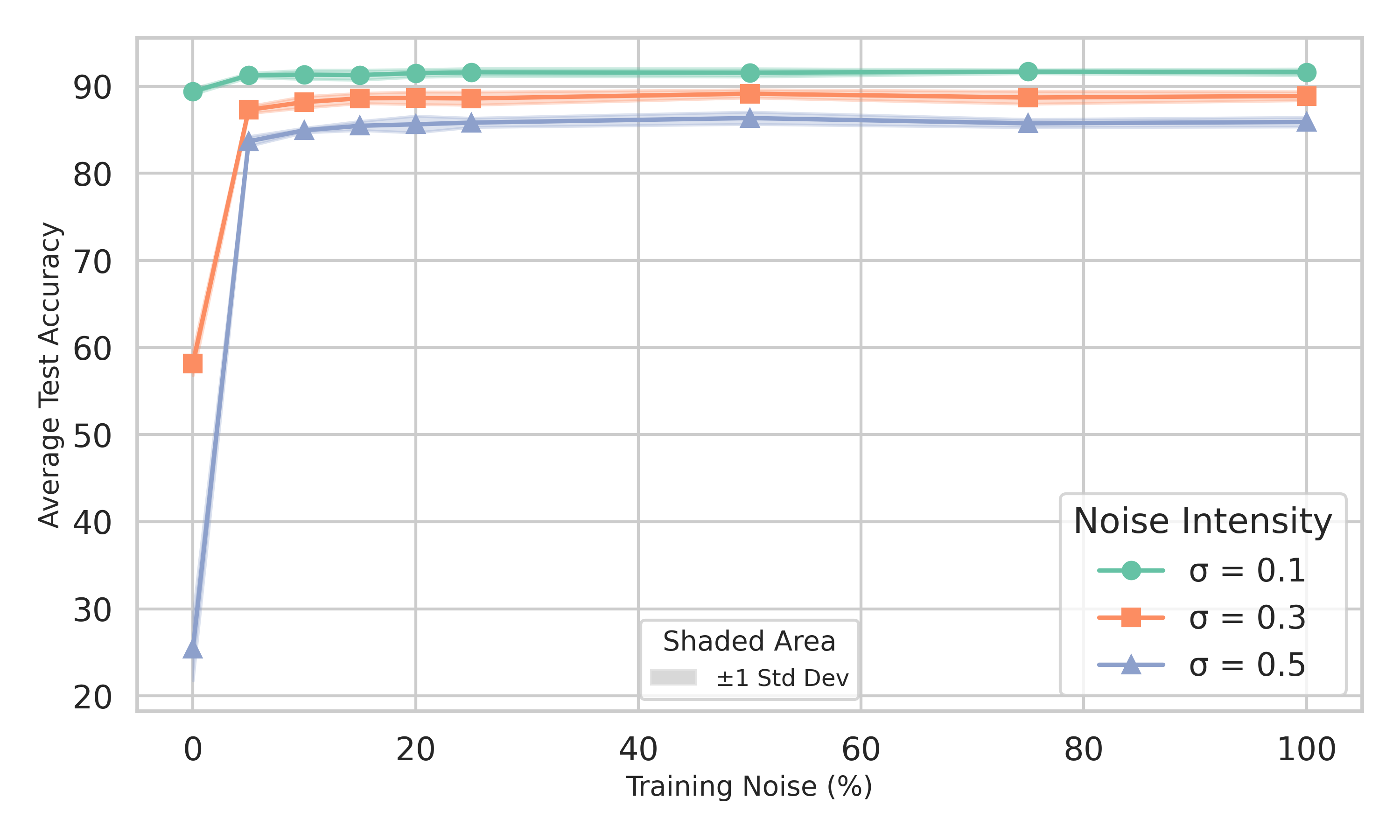}
        \caption{Effect of training noise on fully polluted test performance.}
        \label{fig:acc_gaussian_noisy}
    \end{subfigure}
    \caption{Average test accuracy across different Gaussian noise intensities and varying training noise percentages. Shaded areas indicate $\pm 1$ standard deviation over 10 runs.}
    \label{fig:acc_gaussian_dual}
\end{figure}

Figure~\ref{fig:acc_gaussian_dual} shows the average test accuracy of the model trained on CIFAR-10 data with varying intensities of Gaussian noise. Figure~\ref{fig:acc_gaussian_clean} presents the test performance on the clean test set for these three noise levels, revealing that the model trained with the lowest noise intensity ($\sigma = 0.1$) maintains a relatively stable performance. In contrast, higher noise intensities ($\sigma = 0.3$ and $0.5$) begin to degrade accuracy, particularly as the proportion of noisy training data increases.

Figure~\ref{fig:acc_gaussian_noisy} demonstrates the model’s performance with the respective fully polluted test set. Here, a small amount of noise augmentation during training (as low as $5\%$) leads to a substantial improvement in the accuracy, showing that robustness improves with moderate exposure. Additionally, training with excessively strong noise ($\sigma = 0.5$) leads to a reduction in the average accuracy of the fully polluted (with the respective intensity) test set. Here the results also suggest that moderate  augmentation with low-intensity noise ($\sigma = 0.1$ or $0.3$) achieves the best balance as shown in Figure~\ref{fig:loss_gaussian_dual}. For the strongest intensity ($\sigma = 0.5$), and also considering the results in Sect.~\ref{sec:results_loss_gaussian}, the trained model could be useful for extreme polluted test sets, as it delivers a better performance when compared to the model trained with only clean data.

\subsubsection{Salt-and-Pepper Noise}\label{sec:results_acc_salt_pepper}
\begin{figure}[ht]
    \centering
    \begin{subfigure}[b]{0.48\linewidth}
        \centering
        \includegraphics[width=\linewidth]{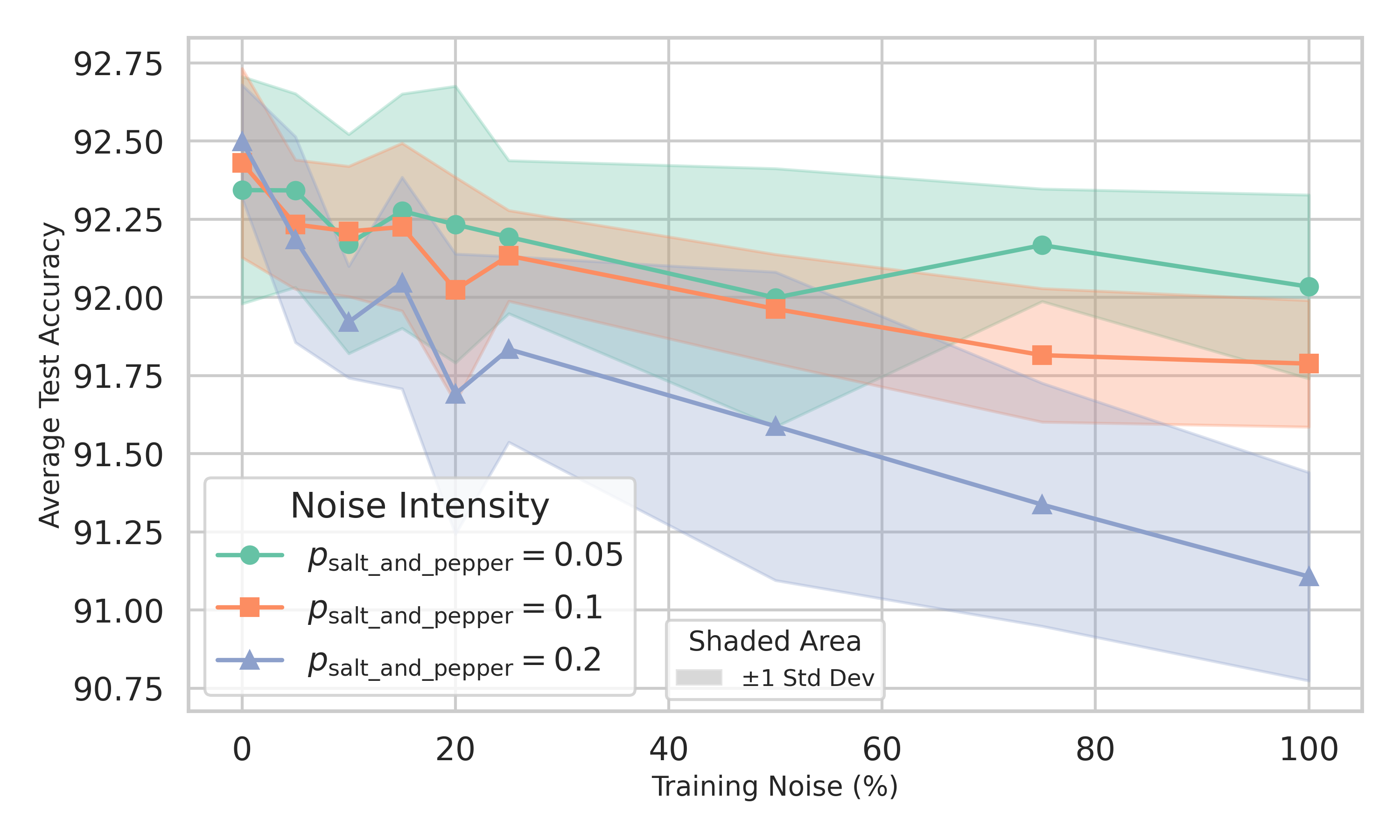}
        \caption{Effect of training noise on fully clean test performance.\\}
        \label{fig:acc_salt_pepper_clean}
    \end{subfigure}
    \hfill
    \begin{subfigure}[b]{0.48\linewidth}
        \centering
        \includegraphics[width=\linewidth]{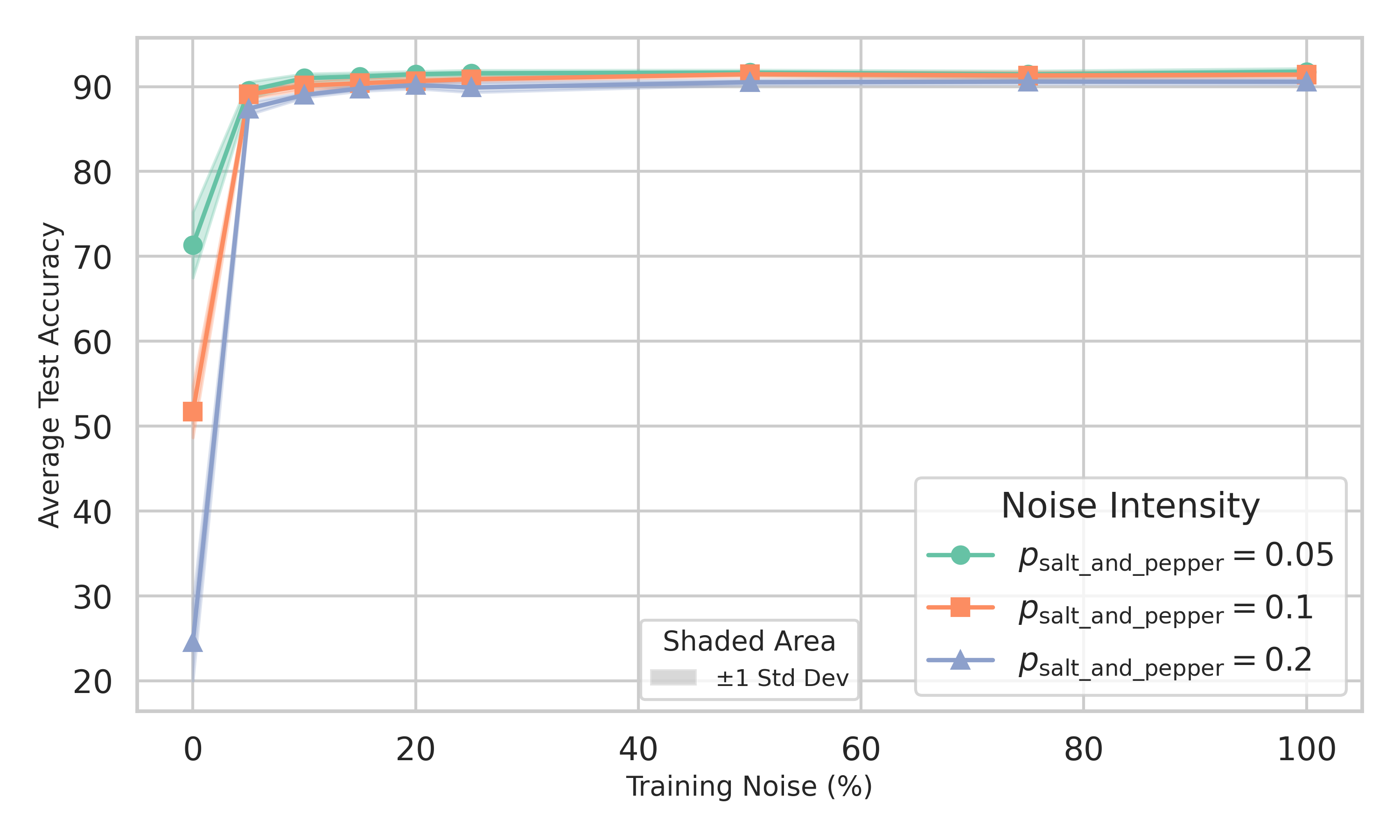}
        \caption{Effect of training noise on fully polluted test performance.}
        \label{fig:acc_salt_pepper_noisy}
    \end{subfigure}
    \caption{Average test accuracy across different Salt-and-Pepper noise intensities and varying training noise percentages. Shaded areas indicate $\pm 1$ standard deviation over 10 runs.}
    \label{fig:acc_salt_pepper_dual}
\end{figure}

Figure~\ref{fig:acc_salt_pepper_dual} shows the impact of Salt-and-Pepper noise across varying noise intensities and proportions of corrupted training data, ranging from 0\% to 100\%. Figure~\ref{fig:acc_salt_pepper_clean} presents the average test accuracy on a clean test set. As the proportion of noisy training data increases, performance on clean data remains relatively stable, with only a slight drop at higher noise intensities. Models trained with low-intensity noise ($p_{\text{salt\_and\_pepper}} = 0.05$) maintain an accuracy around 92\%, while those trained with higher-intensity noise ($p_{\text{salt\_and\_pepper}} = 0.2$) decrease modestly to about 91\%. In contrast, Figure~\ref{fig:acc_salt_pepper_noisy} illustrates performance on a fully corrupted test set, where models trained without noise perform poorly, particularly for $p_{\text{salt\_and\_pepper}} = 0.2$, achieving only about 25\% accuracy. However, introducing even a small proportion of noisy training data (10 to 20\%) leads to a considerable improvement in robustness, with all models reaching approximately 91\% accuracy. These results demonstrate that noise augmentation can substantially enhance robustness to test-time corruption, with minimal impact on clean-data performance.

\subsubsection{Gaussian Blur}\label{sec:results_acc_gaussian_blur}

\begin{figure}[ht]
    \centering
    \begin{subfigure}[b]{0.48\linewidth}
        \centering
        \includegraphics[width=\linewidth]{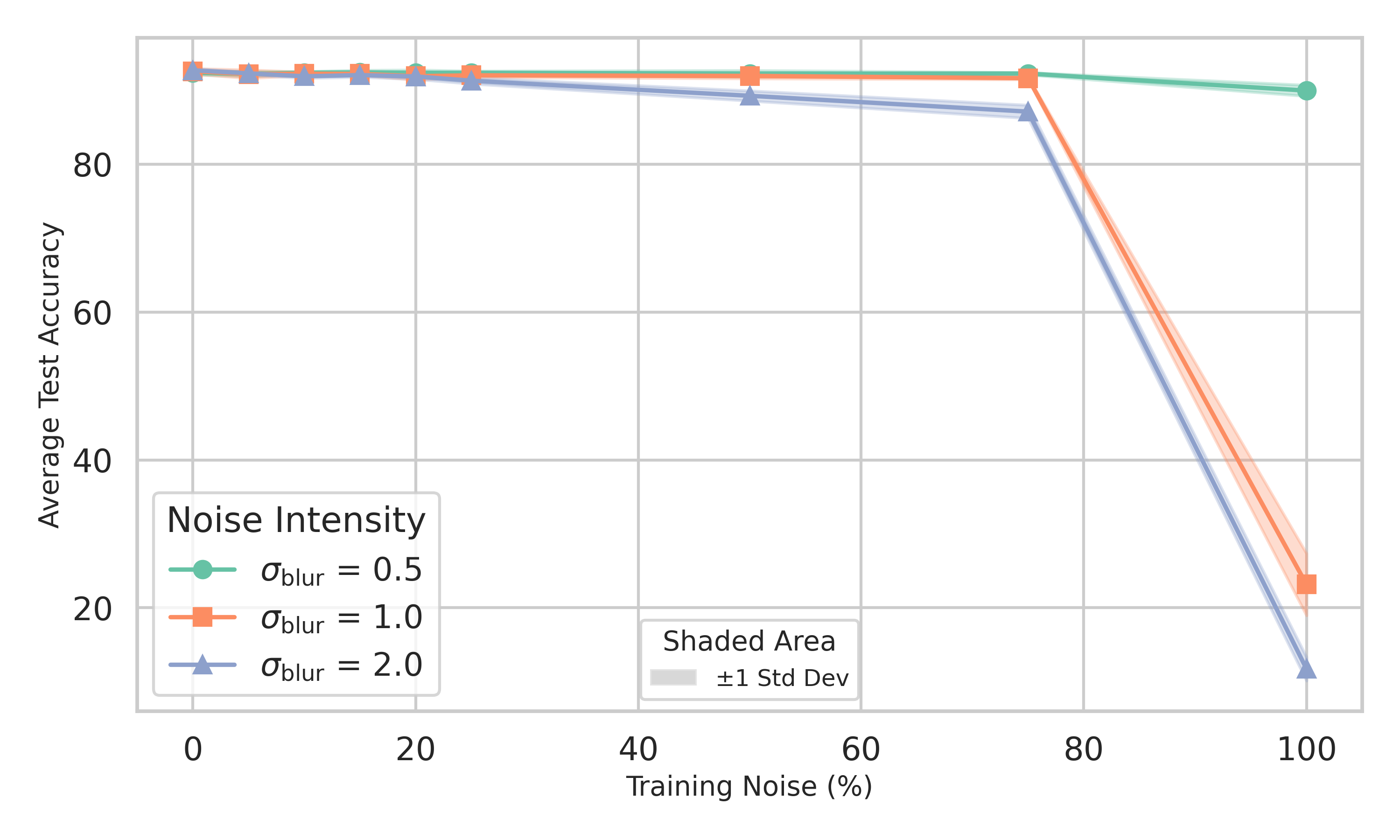}
        \caption{Effect of training noise on fully clean test performance.\\}
        \label{fig:acc_gaussian_blur_clean}
    \end{subfigure}
    \hfill
    \begin{subfigure}[b]{0.48\linewidth}
        \centering
        \includegraphics[width=\linewidth]{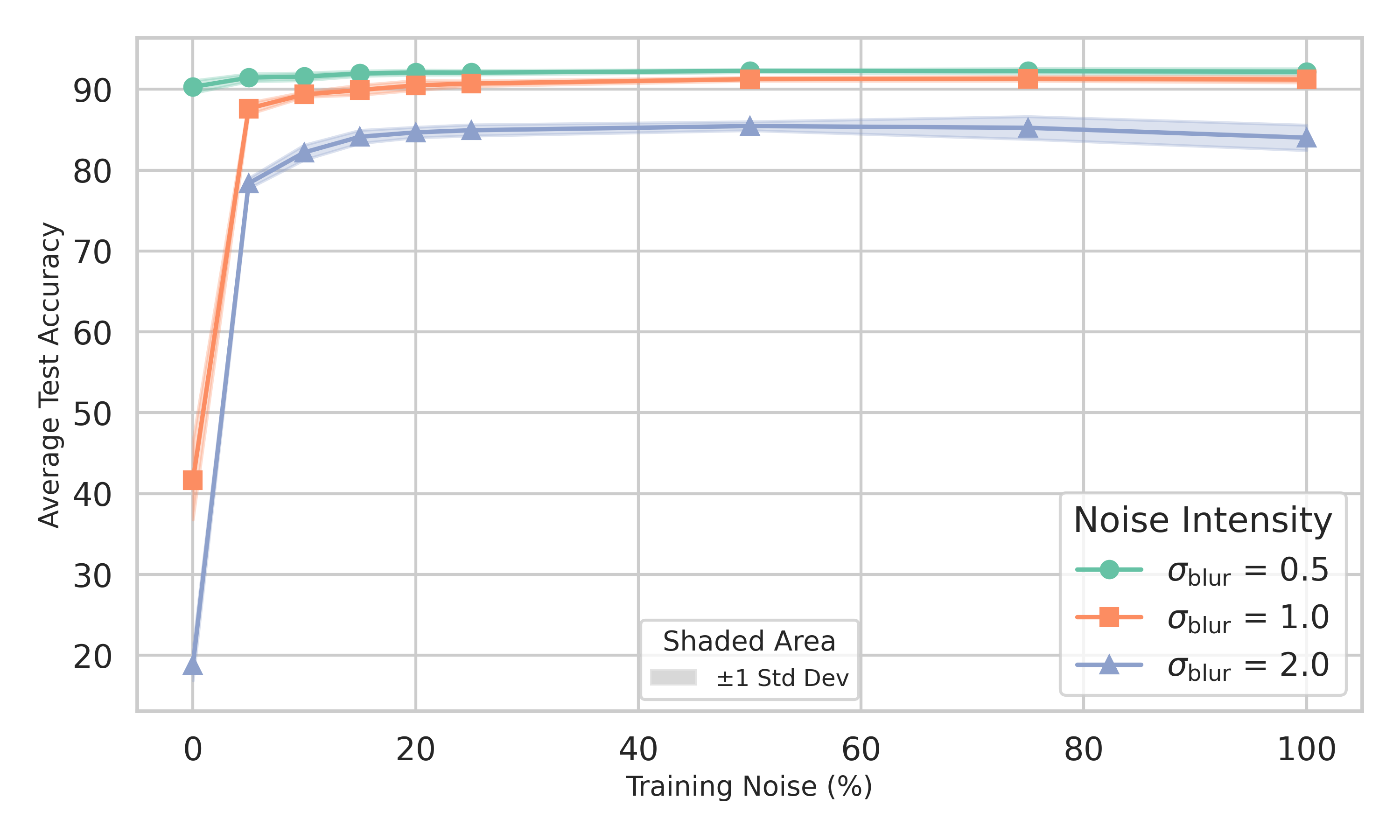}
        \caption{Effect of training noise on fully polluted test performance.}
        \label{fig:acc_gaussian_blur_noisy}
    \end{subfigure}
    \caption{Average test accuracy across different Gaussian blur intensities and varying training noise percentages. Shaded areas indicate $\pm 1$ standard deviation over 10 runs.}
    \label{fig:acc_gaussian_blur_dual}
\end{figure}

Figure~\ref{fig:acc_gaussian_blur_dual} shows the impact of Gaussian blur across varying noise intensities and proportions of corrupted training data, ranging from 0\% to 100\%. Figure~\ref{fig:acc_gaussian_blur_clean} presents the average test accuracy on a clean test set. As the proportion of noisy training data increases, performance on clean data degrades, particularly for higher noise intensities. Models trained with low-intensity blur ($\sigma_{\text{blur}} = 0.5$) maintain relatively stable accuracy around 92\%, while those trained with higher-intensity blur ($\sigma_{\text{blur}} = 1.0$ and $2.0$) drop around 20\%. In contrast, Figure~\ref{fig:acc_gaussian_blur_noisy} illustrates performance on a fully corrupted test set, where models trained without noise perform poorly, especially for $\sigma_{\text{blur}} = 2.0$, achieving only about 20\% accuracy. However, introducing even a small proportion of noisy training data (10 to 20\%) leads to a substantial improvement in robustness, as for moderate levels of blur ($\sigma_{\text{blur}} = 0.5$ and $1.0$) the accuracy reaches approximately 90\% and for the strongest one ($\sigma_{\text{blur}} = 2.0$) the accuracy is above 80\%. These results show that training with the three different levels of blur can significantly enhance robustness.
 
\section{Discussion}\label{sec:discussion}
To analyze the impact of different noise types on model robustness, we examine the average test loss on CIFAR-10 as the training inputs are progressively corrupted. For this analysis, we select the highest tested intensity for each noise type: Gaussian noise ($\sigma = 0.5$), Salt-and-Pepper noise ($p_{\text{salt\_and\_pepper}} = 0.2$), and Gaussian blur ($\sigma_{\text{blur}} = 2.0$).

\subsection{Loss}\label{sec:discussion_loss}

\begin{figure}[ht]
    \centering
    \includegraphics[width=\linewidth]{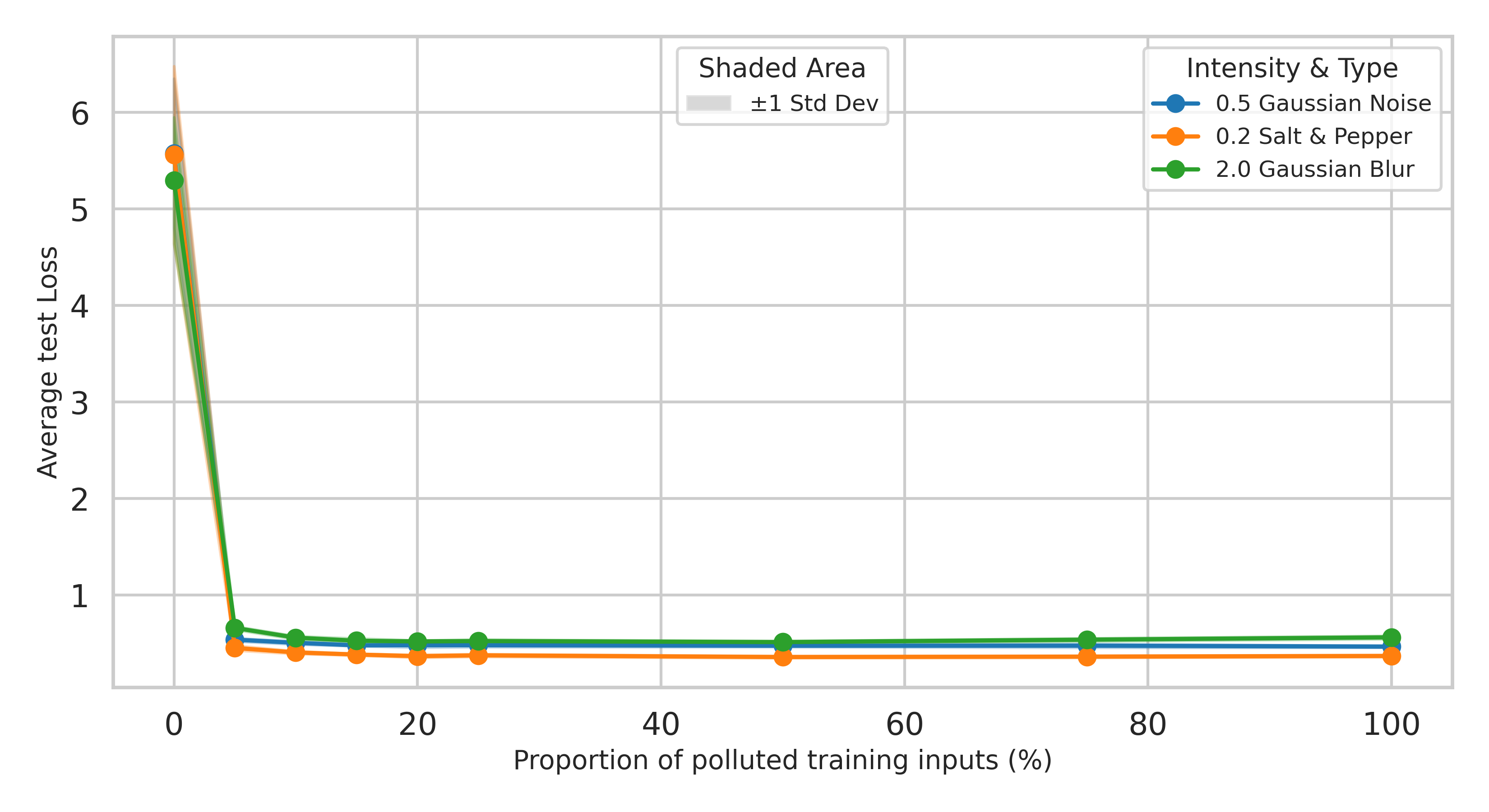}
    \caption{Impact of training noise on all pollution types on highest intensity values in fully polluted testing conditions for the average test Loss.}
    \label{fig:disc_loss}
\end{figure}

As shown in Figure~\ref{fig:disc_loss}, all three noise types, for the highest intensity respectively, exhibit a pronounced drop in test loss with as little as 5\% of the training data being polluted. This early drop suggests that exposure to small amounts of input noise can act as an implicit regularizer, improving generalization. 

Among the three perturbations, Salt-and-Pepper noise consistently yields the lowest test loss across all levels of input corruption, indicating greater robustness to this type of discrete noise. Gaussian noise performs similarly well, whereas Gaussian blur results in slightly higher test loss, particularly at lower corruption levels. However, the differences between the methods diminish beyond 10\% corruption, where all curves flatten, suggesting the model’s ability to adapt to noisy training data once a certain exposure threshold is reached.


\subsection{Accuracy}\label{sec:discussion_acc}

\begin{figure}[ht]
    \centering
    \includegraphics[width=\linewidth]{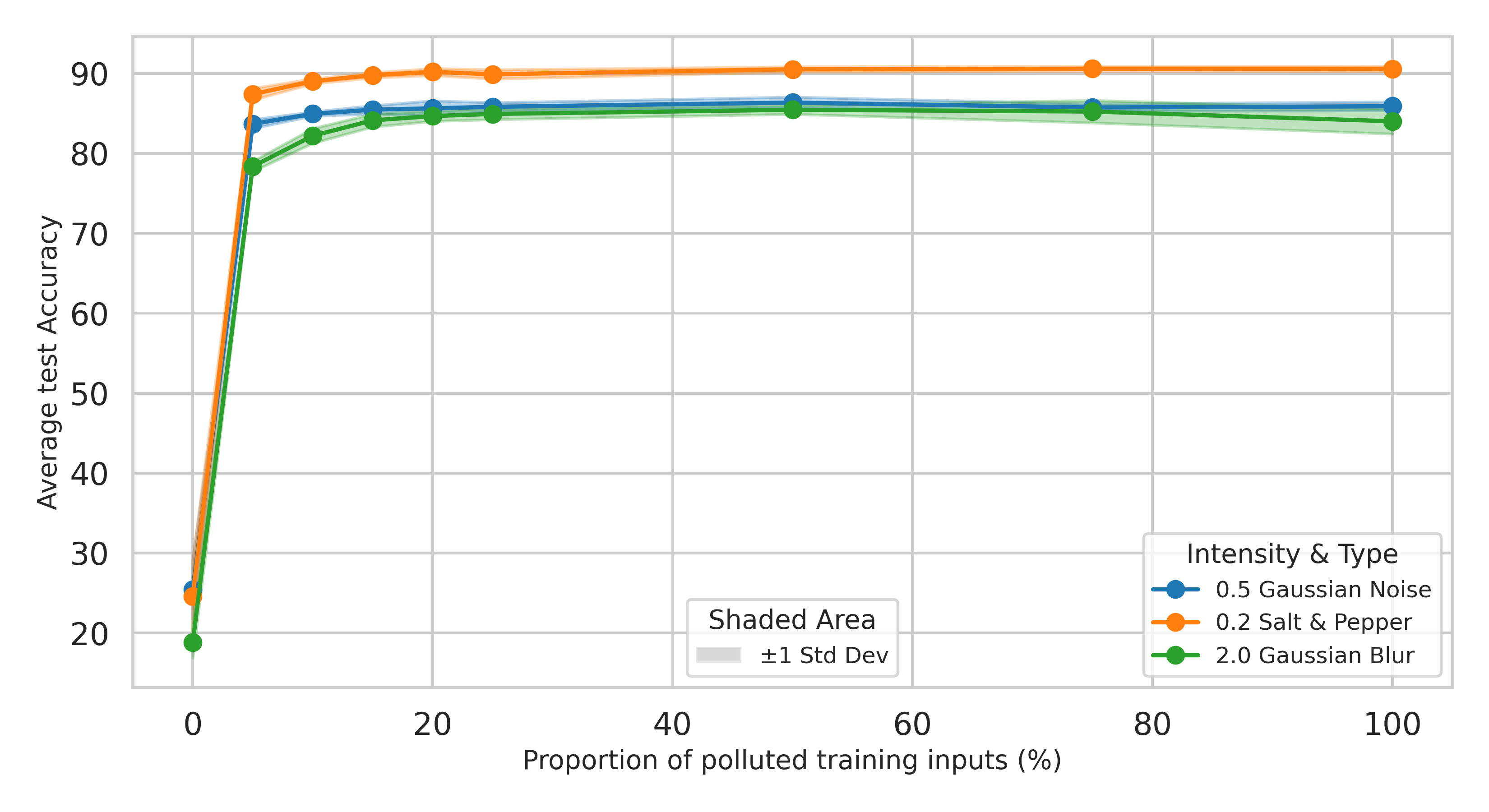}
    \caption{Impact of training noise on all pollution types on highest intensity values in fully polluted testing conditions for the average test accuracy.}
    \label{fig:disc_acc}
\end{figure}

In Figure~\ref{fig:disc_acc} all three noise types exhibit a pronounced enhancement in accuracy loss with as little as 5\% of the training data being polluted. This suggests again that exposure to small amounts of input noise can act as an implicit regularizer, improving generalization. 
Among the noise types, Salt-and-Pepper noise at highest intensity yielded the most robust performance across pollution levels, while Gaussian blur showed the least benefit, particularly at low exposure. Nevertheless, all models exhibited a convergence in performance after about 10\% of noisy training data, highlighting this threshold as sufficient to trigger meaningful robustness without compromising clean-data accuracy.

\subsection{Additional experiments on pollution intensity}\label{sec:discussion_pollution}

Although not reported in the main paper, we are conducting new experiments with stronger intensity values of Salt-and-Pepper noise and Gaussian blur (i.e., $p_{\text{salt\_and\_pepper}} = 0.4$, $\sigma_{\text{blur}} = 4.0$). The preliminary results for loss and accuracy on both the clean and fully polluted test sets remained unchanged. On the clean test set, the loss tends to increase with higher training noise. Accuracy continues to decrease slightly as training noise increases, but the effect is not significant. For the fully polluted test set, the loss is reduced at higher training noise levels, and accuracy improves starting from approximately 5–10\% training noise. Overall, the observed behavior is consistent with the results presented in Section~\ref{sec:results_acc}.

\subsection{Comparison to SOTA}\label{sec:comparison_sota}

We build on prior work that uses data augmentation to improve model robustness. De Vries et al. \cite{devries2017improvedregularizationconvolutionalneural} proposed Cutout, a regularization technique that removes rectangular regions from training images. Although not specifically designed for robustness against corruptions, Cutout improves generalization and enhances clean-data accuracy. Later, Lopes et al. \cite{tradeoffs_clean_noisy} introduced Patch Gaussian augmentation, which injects Gaussian noise into localized patches rather than the entire image. Their method preserves clean-image accuracy while improving robustness to localized corruptions. In contrast, we apply global noise to entire images and evaluate different noise types across a range of intensities and proportions. Our results confirm that even small amounts of noise whether structured or unstructured help CNNs generalize to noisy conditions. While Cutout and Patch Gaussian focus on structural occlusion or local noise, our approach shows that typical sensor degradations like blur and impulse noise can be just as effective for improving robustness. Although our experiments show a degradation in accuracy for higher noise intensities, the findings collectively underscore the value of training with imperfect inputs.

\subsection{Balancing Clean Data and Real-World Robustness}\label{sec:challenge_classical_dq}

This experimental setup indicate that incorporating a degree of noise into the training data can significantly enhance model robustness, especially when test images are subject to similar forms of degradation. These findings suggest that strategically managing data quality by balancing clean and noisy data can lead to more reliable and resilient models suitable for deployment in real-world conditions.

These findings invite a critical reevaluation of classical definitions of data quality, which have traditionally emphasized cleanliness, consistency, and minimal noise. While such criteria aim to optimize model training under ideal conditions, our results suggest that strict adherence to these standards may limit model robustness in real-world scenarios. Introducing controlled amounts of representative noise may slightly deviate from conventional data quality metrics, but it offers clear benefits for generalization to noisy, imperfect inputs. This approach is particularly valuable in safety-critical applications, where operational environments are often unpredictable and robustness is essential.

\section{Conclusion}\label{sec:conclusion}
This study presents an empirical investigation into how the inclusion of noisy data during training influences the robustness of CNNs in image classification tasks. To this end, Gaussian noise, Salt-and-Pepper noise, and Gaussian blur were systematically introduced into varying proportions of the CIFAR-10 training dataset. Model performance was then evaluated on both clean and degraded test inputs. The findings consistently indicate that even limited exposure to noise during training, typically as low as 5\%, can substantially reduce test loss. Notably, incorporating just 10\% noisy data during training significantly enhances robustness, leading to improved accuracy on corrupted test samples.

Overall, the results challenge traditional assumptions regarding the necessity of purely clean datasets for training high-performing models. Instead, they support a more nuanced perspective in which moderate and representative imperfections during training act as a form of regularization, thereby improving model resilience under real-world conditions. These insights highlight the importance of reevaluating data quality criteria in the development of AI systems, particularly for safety-critical or real-world applications where input imperfections are the norm.

\section{Acknowledgments}
I would like to thank the Data Quality team at the DLR Institute for AI Safety and Security for their valuable feedback. In particular, I am grateful to Leonie Etzold for the constructive discussions, which have significantly contributed to improving this work.

\appendix    

\section{CIFAR-10}\label{sec:cifar-10}

\subsection{Loss clean dataset}
\begin{figure}[h!]
 \centering
 \includegraphics[scale=0.40]{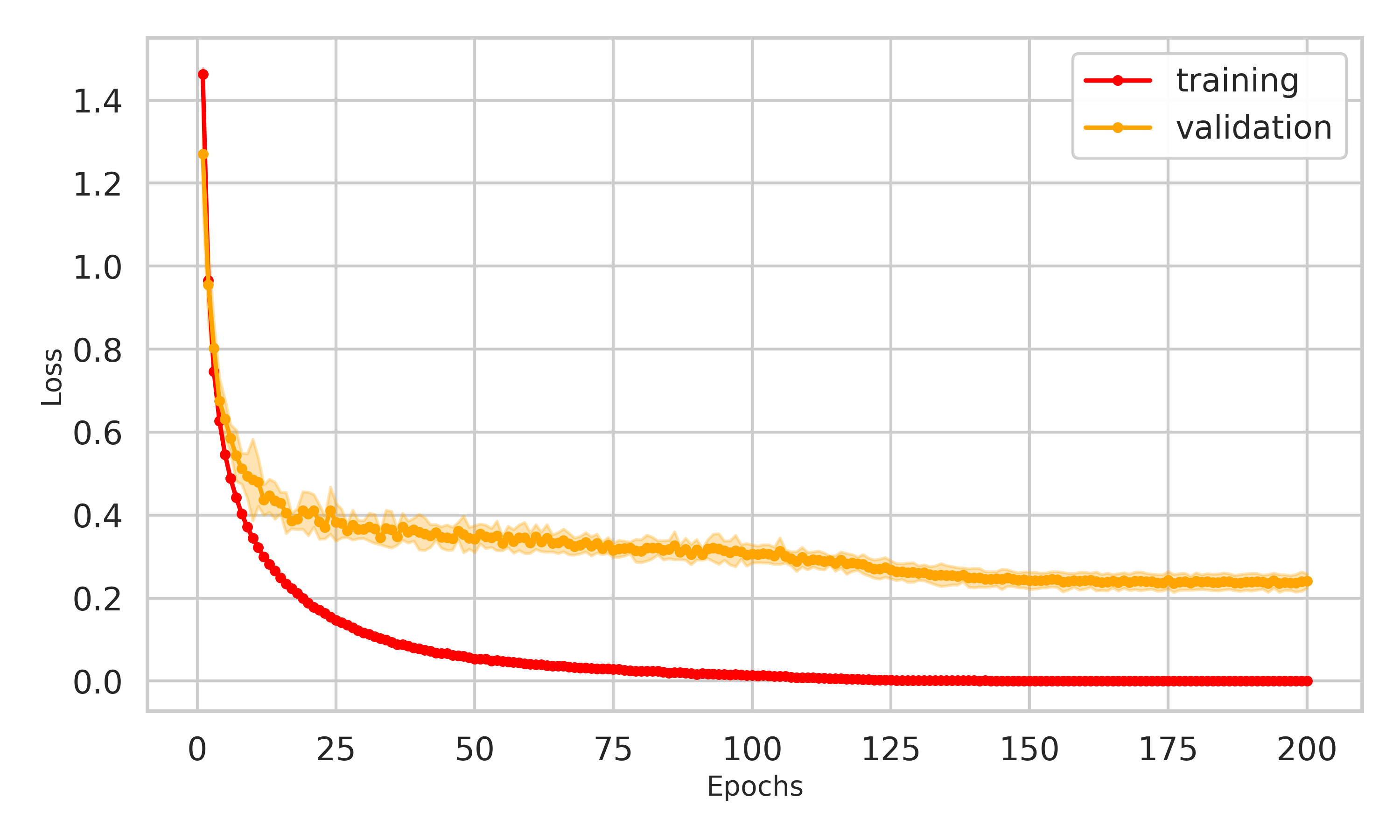}
 \caption{Loss single clean dataset.}
 \label{fig:loss_clean_runs10}
\end{figure}

To assess the impact of non-determinism during training, we trained the model on CIFAR-10 over 10 independent runs and report the mean and standard deviation of the training and validation loss across 200 epochs (see Figure~\ref{fig:loss_clean_runs10}). The training loss consistently decreases across all runs, exhibiting a smooth exponential decay with negligible variance, and converges near zero. This indicates that the model is able to fit the training data well, and that training randomness, such as random initialization, data shuffling, and dropout, has minimal impact on the convergence of the training loss.

In contrast, the validation loss decreases rapidly during the initial 20–30 epochs, then continues to decrease at a slower rate before plateauing around epoch 100. The final validation loss stabilizes around 0.25 with a noticeably higher variance compared to the training loss, especially in the early and intermediate epochs. This behavior reflects the model's generalization capacity and highlights a moderate sensitivity to non-deterministic factors, which primarily affect validation performance rather than training convergence. The increasing gap between training and validation loss over time suggests a degree of overfitting, which is typical for deep models trained on CIFAR-10 without aggressive regularization. These results illustrate that while the optimization dynamics are largely robust to stochastic effects, the model's generalization is more susceptible to variability introduced by non-determinism in the training pipeline.

\subsection{Accuracy clean dataset}

\begin{figure}[h!]
 \centering
 \includegraphics[scale=0.40]{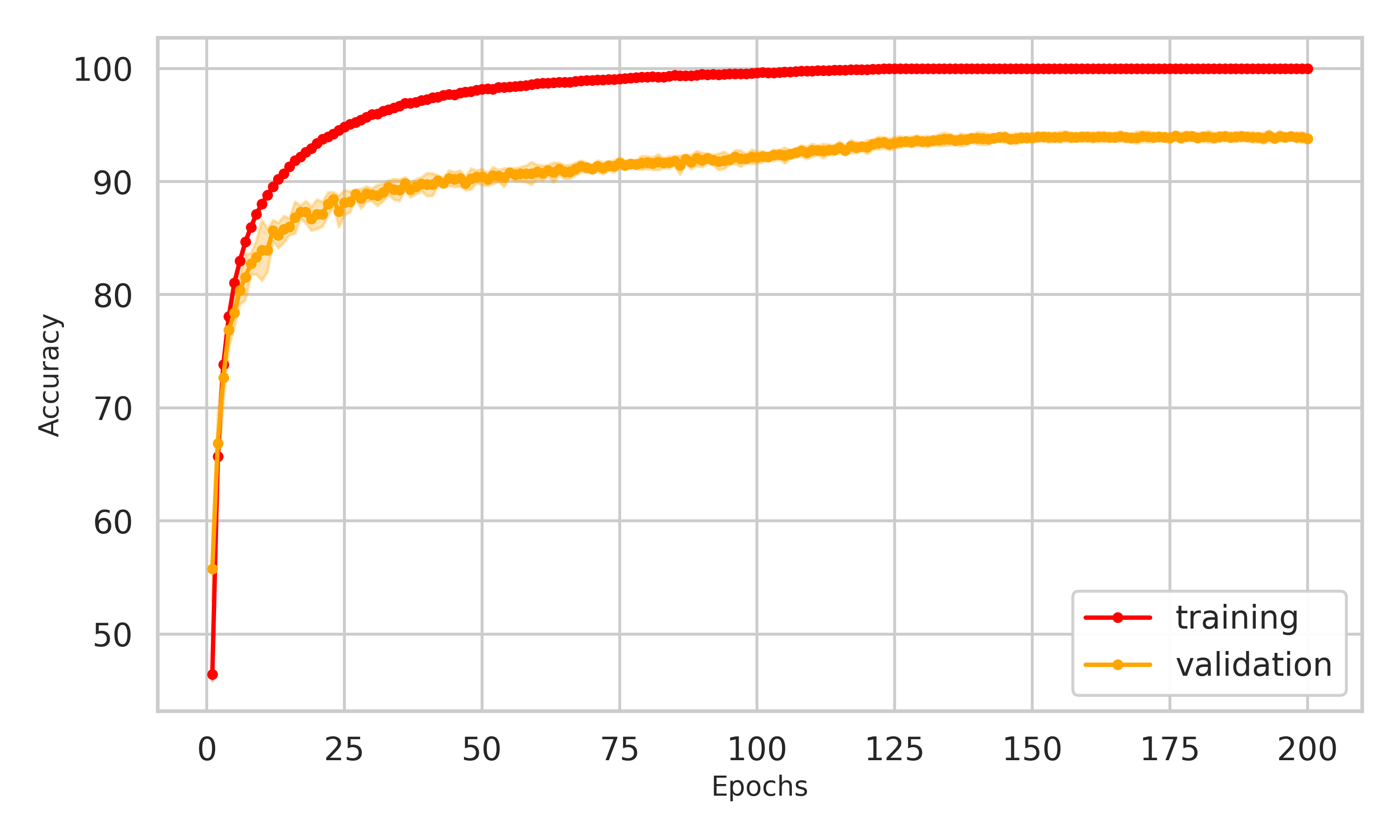}
 \caption{Accuracy single clean dataset.}
 \label{fig:acc_clean_runs10}
\end{figure}

Figure~\ref{fig:acc_clean_runs10} illustrates the training and validation accuracy over 200 epochs, averaged across 10 independent runs, with shaded regions indicating the standard deviation. The training accuracy exhibits a steady increase throughout training, rapidly approaching near-perfect performance ($>$99\%) and plateauing after approximately 100 epochs. This reflects the model's strong capacity to fit the CIFAR-10 training data, with minimal variability across runs, confirming the robustness of the optimization process under stochastic conditions.

The validation accuracy also improves sharply during the early epochs, reaching approximately 90\% within the first 30 epochs, and continues to increase more gradually thereafter. It ultimately stabilizes around 93\%, with a standard deviation slightly larger than that of the training accuracy. This suggests that while the model generalizes well, its performance on unseen data is more sensitive to random factors such as initialization and data ordering. The persistent gap between training and validation accuracy indicates mild overfitting, consistent with the trends observed in the loss curves. Nonetheless, the relatively narrow variance bands demonstrate that the generalization behavior remains stable across repeated runs, highlighting that while non-determinism introduces some variability, it does not significantly undermine the model's overall performance on the validation set. 

Finally, Table~\ref{tab:test_accuracy} shows the final test accuracy for the 10 different classes of CIFAR-10 for 100 and 200 epochs, respectively. Given the marginal improvement between the two configurations, we fixed the number of training epochs to 100 epochs for the experiments presented in this paper.  

\begin{table}[ht]
\centering
\caption{Final classification accuracy per class on the CIFAR-10 for 100 and 200 epcohs.}\label{tab:test_accuracy}
\begin{tabular}{l c c } 
\toprule
\textbf{Class} & \textbf{Accuracy (\%)}& \textbf{Accuracy (\%)} \\ 
               &    (100 epochs)       &    (200 epochs) \\ 
\midrule
Plane  &  94.30 & 95.60 \\ 
Car    &  96.50 & 98.10 \\ 
Bird   &  89.80 & 91.40 \\ 
Cat    &  83.70 & 87.70 \\ 
Deer   &  93.80 & 95.00 \\ 
Dog    &  88.60 & 89.60 \\ 
Frog   &  95.60 & 96.40 \\ 
Horse  &  95.80 & 95.80 \\ 
Ship   &  95.00 & 96.60 \\ 
Truck  &  95.80 & 96.00 \\ 
\midrule 
\textbf{Final test accuracy} &  \textbf{92.89} & \textbf{94.22} \\ 
\bottomrule
\end{tabular}
\end{table}

\pagebreak

\bibliography{report} 
\bibliographystyle{spiebib} 

\end{document}